\documentclass[iicol]{sn-jnl}
\usepackage{etex}

\jyear{2022}%

\usepackage{amssymb}
\usepackage{amsthm}
\usepackage{amsmath}
\theoremstyle{lemma}
\theoremstyle{proposition}
\newtheorem{lemma}{Lemma}
\newtheorem{proposition}{Proposition}
\theoremstyle{definition}
\theoremstyle{theorem}
\newtheorem{theorem}{Theorem}

\usepackage{tabularx}
\usepackage{algpseudocode}

\usepackage[footnotesize]{caption}
\usepackage{graphicx}
\usepackage{subfigure}
\usepackage{enumitem}
\renewcommand\theenumi{\alph{enumi}}
\usepackage{url}
\usepackage{multirow}
\usepackage{booktabs}
\usepackage{lineno}
\usepackage{stfloats}
\usepackage{enumerate}
\usepackage{xcolor}
\definecolor{lightpink}{HTML}{FFB6C1}
\definecolor{lightskyblue}{HTML}{87CEFA}
\begin{document}
\title[Article Title]{Fast Newton method solving KLR based on Multilevel Circulant Matrix with log-linear complexity}


\author[1]{\fnm{Junna} \sur{Zhang}}\email{junnazhang@stu.xidian.edu.cn}

\author*[1]{\fnm{Shuisheng} \sur{Zhou}}\email{sszhou@mail.xidian.edu.cn}

\author[1]{\fnm{Cui} \sur{Fu}}\email{cuifu@stu.xidian.edu.cn}
\author[1]{\fnm{Feng} \sur{Ye}}\email{fye@xidian.edu.cn}

\affil*[1]{\orgdiv{School of Mathematics and Statistics}, \orgname{Xidian University}, \orgaddress{\city{Xi'an}, \postcode{710126}, \country{China}}}


\abstract{Kernel logistic regression (KLR) is a conventional nonlinear classifier in machine learning. With the explosive growth of data size, the storage and computation of large dense kernel matrices is a major challenge in scaling KLR. Even the nystr\"{o}m approximation is applied to solve KLR, it also faces the time complexity of $O(nc^2)$ and the space complexity of $O(nc)$, where $n$ is the number of training instances and $c$ is the sampling size. In this paper, we propose a fast  Newton method  efficiently solving large-scale KLR problems by exploiting the storage and computing advantages of multilevel circulant matrix (MCM). Specifically, by approximating the kernel matrix with an MCM, the storage space is reduced to $O(n)$, and further approximating the coefficient matrix of the Newton equation as MCM, the computational complexity of Newton iteration is reduced to $O(n \log n)$. The proposed method can run in log-linear time complexity per iteration, because the multiplication of MCM (or its inverse) and vector can be implemented the multidimensional fast Fourier transform (mFFT). Experimental results on some large-scale binary-classification and multi-classification problems show that the proposed method enables KLR to scale to large scale problems with less memory consumption and less training time without sacrificing test accuracy.}

\keywords{Kernel logistic regression, Newton method, Large scale, Multilevel circulant matrix approximation}



\maketitle


\section{Introduction}\label{1}
Kernel logistic regression (KLR) is a log-linear model with direct probabilistic interpretation and can be naturally extended to multi-class classification problems \cite{jaakkola1999probabilistic}. It is widely used in many fields, including automatic disease diagnosis \cite{choudhury2021predicting}, detecting fraud \cite{TJLT201908008}, landslide susceptibility mapping \cite{hong2015spatial}, etc. But it is difficult to scale to large-scale problems due to its high space and time complexity. A key issue in extending KLR to large-scale problems is the storage and computation of the kernel matrix, which is usually intensive. In particular, when Newton method is used to solve KLR, the inverse operation of the Hessian matrix in each iteration requires $O(n^3)$  time and requires $O(n^2)$ space to store the kernel matrix.

A number of researchers have been working to make KLR feasible for large-scale problems \cite{zhu2005kernel, sugiyama2010computationally, keerthi2005fast}. They mainly start from the following two aspects: sparsity of solutions or decomposing the original problem into subproblems. Inspired by the sparsity of support vector machine (SVM), Zhu and Hastie \cite{zhu2005kernel} proposed the import vector machine (IVM) algorithm to reduced the time complexity of the binary classification of KLR to $O(n^{2} s^{2})$, where $s$ is the number of import points. IVM is still difficult to calculate for large-scale problems although the complexity has been reduced. Inspired by sequence minimum optimization algorithm (SMO) solving SVM, Keerthi et al. \cite{keerthi2005fast} proposed a fast dual algorithm for solving KLR. By continuously decomposing the original problem into subproblems, the fast dual algorithm only updates two variables per iteration. However, the time cost of iteratively updating values in the fast dual algorithm is increased by the introduction of the kernel matrix calculation. In short, existing methods still make it difficult to scale KLR to large-scale problems.

In this paper, we focus on kernel approximation to accelerate large-scale KLR inspired by the excellent performance of kernel approximation in learning problems \cite{NIPS2000_19de10ad, lei2020improved, jia2020large,chen2021fast}. A great deal of work has been done on kernel approximation. The commonly used kernel approximation methods include Nystr{\"o}m method \cite{li2014large, calandriello2016analysis, NIPS2015_f3f27a32, NIPS2017_a03fa308, kumar2012sampling, gittens2016revisiting}, random feature \cite{NIPS2007_013a006f,he2019fast, pmlr-v28-le13, feng2015random, xiong8959408kui, dao2017gaussian, munkhoeva2018quadrature, Zhuli2021towards, liu2020random}, multilevel circulant matrix (MCM) \cite{song2009approximation, song2010approximation, ding2011approximate,edwards2013approximate, ding2017approximate, ding2020approximate}, and so on.

Nystr{\"o}m method is the classical kernel matrix approximation, whose outstanding feature is the sampling of data before large-scale matrix operations. After the Nystr{\"o}m method successfully and efficiently solves large-scale Gaussian processes \cite{NIPS2000_19de10ad}, a number of sampling methods with strong theoretical guarantees have been proposed to satisfy the desired approximation with fewer sampling points. Among these sampling methods, the leverage score sampling technique \cite{NIPS2015_f3f27a32} is the most widely used in practical applications. Recursive ridge leverage scores (RRLS) \cite {NIPS2017_a03fa308} finds a more accurate kernel approximation in less time by employing a fast recursive sampling scheme. However, in \cite{yin2019sketch} the authors pointed out that the Nystr{\"o}m approximation is very sensitive to inhomogeneities in the sample.

The random feature is constructed randomly from a nonlinear mapping of the input space to the Hilbert space, that is, the direct approximation of the kernel function without calculating the elements in the kernel matrix. Rahimi and Recht \cite{NIPS2007_013a006f} proposed the random feature map of shifted invariant kernel functions based on Fourier transform. In order to speed up feature projection, Feng et al. \cite{feng2015random} proposed structured random matrices, signed Circulant Random Matrix (CRM), to project input data. The feature mapping can be done in $O(n D \log d)$  time by using the fast Fourier Transform (FFT), where $d$ and $D$ represent the dimensions of the input data and the random feature space, respectively. Li et al. \cite{Zhuli2021towards} provided the first unified risk analysis of learning with random Fourier features and proposed leverage score random feature mapping which  needs $O(nD^2+D^3)$ time to generate refined random features. Obviously, when $d$ or $D$  is very large, the feature mapping is costly.

The idea of MCM approximate kernel matrix was first proposed by Song and Xu \cite{song2010approximation} and many theoretical results are proven.
By approximated the kernel matrix with MCM, researchers have developed many applications in different machine learning areas, such as  the kernel ridge regression \cite{edwards2013approximate}, automatic kernel selection problem \cite{ding2017approximate} and least squares support vector machines \cite{ding2020approximate}, where the approximated kernel matrix is stored in $O(n)$ and the computational complexity of the corresponding algorithms is only $O(n \log n)$.
Since MCM can save a lot of memory and has certain computing advantages, we choose MCM approximation to speed up KLR.

In the works \cite{edwards2013approximate,ding2017approximate,ding2020approximate}, the core problem is to solve a system of linear equations $(\boldsymbol K + n\lambda \boldsymbol I){\boldsymbol {d}}={\boldsymbol {b}}$, where $\boldsymbol K$ is a kernel matrix or the linear combination of multiple kernel matrices and $\lambda$ is the regularized parameter. If $\boldsymbol K$ is approximated by an MCM, then $(\boldsymbol K + n\lambda \boldsymbol I)$ is MCM too, hence the system of linear equations can be solved in $O(n \log n)$ time by the multidimensional fast Fourier transform (mFFT) owing to the advantages of MCM.
When applying MCM directly to KLR, it faces to solve a system of linear equations $\left({\boldsymbol K}^ \top\boldsymbol \Lambda \boldsymbol K + n\lambda \boldsymbol K\right)\boldsymbol d=\boldsymbol b.$  Only approximated $\boldsymbol K$ as an MCM, the coefficient matrix of the system of linear equations is still not an MCM. Hence it cannot be solved in $O(n \log n)$ and still suffers from high computational complexity. In this case, the most efficient scheme to solve it is to run conjugate gradient method $T$ loops with the computational complexity $O(Tn\log n)$. Since $T=O(n)$ for conjugate gradient method \cite{hestenes1952methods}, this scheme is still insufferable for large-scale problems.

In this work, to effectively solve KLR with large-scale training samples, we firstly simplify the resulted Newton equation, then approximate the kernel matrix by MCM as \cite{edwards2013approximate, ding2017approximate,ding2020approximate} did. Further  we approximate the coefficient matrix of the simplified Newton equation as an MCM too. Hence, we propose a fast Newton method based on MCM which can efficient solve large-scale KLR with $O(n)$ space complexity and $O(n\log n)$ computational complexity. Many experimental results support that the proposed method can make KLR problem scalable.

The rest of the paper is organized as follows. In Section \ref{section:2}, we review KLR and MCM. In Section \ref{section:3}, we present the fast Newton method based on MCM approximation for solving KLR. We report experimental results in  Section \ref{section:4}. Section \ref{section:5} concludes this paper.
\section{Preliminaries}
\label{section:2}
In this section, we review KLR and MCM and introduce some interesting properties of MCM.
\subsection{Kernel Logistic Regression}
\label{subsection:2.1}
Given the training set $\mathbb{D}=\{(\boldsymbol{x}_{i}, \boldsymbol y_{i}), i=1,\ldots,n\}$, where $\boldsymbol{x}_{i} \in \mathcal{X} \in \mathbb{R}^{d}$ is the input data and $\boldsymbol  y_{i} \in\{0,1\}$ is the output targets corresponding to the input. A reproducing kernel Hibert space $\mathbb{H}$ is defined by a kernel function $\kappa(\boldsymbol x, \boldsymbol z)=\left< \varphi(\boldsymbol x), \varphi(\boldsymbol z) \right>$ with $\varphi:\mathbb{R}^{d}\mapsto\mathbb{H}$,  which measures the inner product between the input vectors in the feature space. Then a traditional logistic regression model \cite{cawley2004efficient} is constructed in the feature space as follows
\begin{equation}\label{equ:1}
\min _{\boldsymbol w \in \mathbb{H}}\frac{\lambda}{2} {\Vert \boldsymbol w \Vert}^2-\frac{1}{n}(\boldsymbol y^{ \top } \ln \boldsymbol p+(\boldsymbol 1-\boldsymbol y)^{ \top } \ln (\boldsymbol 1-\boldsymbol p)),
\end{equation}
where $\boldsymbol  p_{i}=1/(1+e^{-\left< \boldsymbol w, \varphi(\boldsymbol x_{i})\right>})$ is the posterior probability estimation of $\boldsymbol  y_{i}=1$, $\lambda>0$  is the regularization parameter, and $\boldsymbol {1}$ is the all-one vector.

By the representer theorem \cite{scholkopf2001generalized}, the solution to the optimization problem \eqref {equ:1} can be represented as
\begin{equation}\label{equ:2}
\boldsymbol w=\sum_{i=1}^{n} \boldsymbol \alpha_{i} \varphi\left(\boldsymbol{x}_{i}\right),
\end{equation}
where $\boldsymbol \alpha \in \mathbb{R}^{n}$. Then plugging \eqref {equ:2} in \eqref {equ:1}, we can get the following KLR model
\begin{equation}\label{equ:3}
\mathop {\min }\limits_{\boldsymbol \alpha \!\in {\mathbb{R}^n}} \!F(\boldsymbol \alpha )\!=\!\frac{\lambda }
{2}{\boldsymbol \alpha \!^\top}\boldsymbol K\!\boldsymbol \alpha \!-\! \frac{1}
{n}({\boldsymbol y\!^\top}\!\ln \boldsymbol p \!+\! {{(\boldsymbol 1 \!-\! \boldsymbol y)}\!^\top}\!\ln (\boldsymbol 1 \!-\! \boldsymbol p)),
\end{equation}
where $\boldsymbol K$  is the kernel matrix satisfying $\boldsymbol K_{i, j}=\kappa(\boldsymbol x_{i}, \boldsymbol x_{j})$, $ {\boldsymbol p}_{i}=1/({{1 + {e^{ - \boldsymbol K(i, \cdot)\boldsymbol \alpha }}}})$, and $\boldsymbol K(i, \cdot)$ denotes the  $i$-th row of the kernel matrix.

KLR is a convex optimization problem, which can be solved by Newton method \cite{dennis1996numerical} with quadratic convergence rate. However, Newton method requires $O(n^{3})$ time complexity and $O(n^{2})$ space complexity for each iteration, which is not feasible for large-scale data sets. Therefore, we need a more effective method to solve KLR.

\subsection{ Multilevel Circulant Matrix Approximation }
\label{subsection:2.2}
Here we introduce the concept of MCM and some of its interesting properties, and analyze its computational advantages.

To facilitate representation, we introduce the notion of multilevel indexing  \cite{song2010approximation}. In order to construct a $q$-level circulant matrix of level order $\boldsymbol{q}$, it is necessary to decompose $n\in \mathbb{N}$ into the product of $q\in \mathbb{N}$ positive integers, that is, $n=n_{0} n_{1} \cdots n_{q-1}$. We denote $\boldsymbol{q}:=\left[n_{0}, n_{1},\ldots, n_{q-1}\right] \in \mathbb{N}^{q}$. Then, multilevel indexing $[\boldsymbol{q}]$ of the $q$-level circulant matrix is defined as follows
\begin{equation}
[\boldsymbol{q}]\!:=\!\left[n_{0}\right]\!\times\!\left[n_{1}\right]\!\times\!\cdots\!\times\![n_{q-1}]\!\in \mathbb{R}^{n \times q}\!,\tag{Cartesian product}
\end{equation}
where $[n_{q-j}]:=\{0,1,\ldots, n_{q-j}\},j=1,2,\ldots,q$.

According to \cite{tyrtyshnikov1996unifying}, if a matrix $\boldsymbol K_{\boldsymbol{q}}$ consists of $n_0 \times n_0$   blocks and each block $(q-1)$-level circulant matrix of level order $[n_1,\ldots, n_{q-1}]$, then $\boldsymbol K_{\boldsymbol{q}}$ is called a  $q$-level circulant matrix. In other words, an MCM is a matrix that can be partitioned into blocks, which are further partitioned into smaller blocks. Specifically, $\boldsymbol K_{\boldsymbol{q}}= [\boldsymbol K_{\boldsymbol {i},\boldsymbol {j}}:\boldsymbol {i},\boldsymbol {j} \in  [\boldsymbol{q}]]$ is called a $q$-level circulant matrix if for any $\boldsymbol {i}=({\boldsymbol i}_{0},{\boldsymbol i}_{1},\ldots,{\boldsymbol i}_{q-1}) \in [\boldsymbol{q}]$, $\boldsymbol {j}=({\boldsymbol j}_{0},{\boldsymbol j}_{1},\ldots,{\boldsymbol j}_{q-1}) \in [\boldsymbol{q}]$,
\begin{equation*}
\boldsymbol K_{\boldsymbol i,\boldsymbol j}=\boldsymbol k_{{\boldsymbol i}_{0}-\bmod \left({\boldsymbol j}_{0}, n_{0}\right),\ldots, {\boldsymbol i}_{q-1}-\bmod ({\boldsymbol j}_{q-1}, n_{q-1})},
\end{equation*}
where $\boldsymbol k$  is the first column of $\boldsymbol K_{\boldsymbol{q}}$. Then a $q$-level circulant matrix $\boldsymbol K_{\boldsymbol{q}}$ is fully determined by its first column.
So we write ${\boldsymbol K_{\boldsymbol{q}}} = {circ_{\boldsymbol{q}}}[{\boldsymbol k}]$.

The computational advantages of MCM are analyzed in detail in \cite{davis1979circulant}, and the key conclusions are restated as follows.
\begin{lemma}\label{lem:1}\cite{davis1979circulant}
Suppose that $\boldsymbol K_{\boldsymbol{q}}$  is an MCM of level order $\boldsymbol{q}$  and $\boldsymbol k$  is its first column. Then  $\boldsymbol K_{\boldsymbol{q}}$  is a $q$-level circulant matrix of level order $\boldsymbol{q}$   if and only if
\begin{equation}\label {equ:4}
\boldsymbol K_{\boldsymbol{q}}=\frac{1}{n} \boldsymbol \phi^{*} \operatorname{diag}(\boldsymbol \phi \boldsymbol k) \boldsymbol \phi,
\end{equation}
where $\boldsymbol \phi=\boldsymbol F_{n_{0}} \otimes \boldsymbol F_{n_{1}} \otimes \cdots \otimes \boldsymbol F_{n_{q-1}}$,  $\boldsymbol A\otimes \boldsymbol B$ denotes the Kronecker product of  $\boldsymbol A$ and $\boldsymbol B$,  and  $\boldsymbol F_{n_{q-j}}=[e^{(2 \pi i / n_{q-j}) st}: s, t \in[n_{q-j}]]$, $j=1,2,\ldots, q$  with  $i$ being the imaginary unit.
\end{lemma}

\begin{theorem}\label{the0:1}\cite{davis1979circulant}
Assume that $\boldsymbol A_{\boldsymbol{q}}$  and $\boldsymbol B_{\boldsymbol{q}}$ are both MCM of level order $\boldsymbol{q}$, then $\boldsymbol A_{\boldsymbol{q}}+\boldsymbol B_{\boldsymbol{q}}$  is also an MCM of level order  $\boldsymbol{q}$.
\end{theorem}

\begin{theorem}\label{the0:2}\cite{davis1979circulant}
Assume that $\boldsymbol A_{\boldsymbol{q}}$  is an invertible MCM of level order $\boldsymbol{q}$  and $\boldsymbol a$  is the first column of $\boldsymbol A_{\boldsymbol{q}}$, $\boldsymbol  \nu=\boldsymbol \phi \boldsymbol a$ is the vector of eigenvalues, then $\boldsymbol A_{\boldsymbol{q}}^{ - 1}$ is also an MCM, and $\boldsymbol A_{\boldsymbol{q}}^{ - 1}=(1/n) \boldsymbol \phi^{*} (\operatorname{diag}({\boldsymbol \nu}))^{-1} \boldsymbol \phi $.
\end{theorem}

The following Algorithm \ref {alg:1} proposed in \cite{song2009approximation} can construct an MCM $\boldsymbol K_{\boldsymbol{q}}$ from a kernel function to approximate the kernel matrix $\boldsymbol K$.

\begin{algorithm}[H]
\caption{Construction of an MCM \cite{song2009approximation}}\label{alg:1}
\begin{algorithmic}[1]
\Require a kernel function $\kappa$, a sequence of positive numbers $\boldsymbol {h} = ({\boldsymbol {h}_0},{\boldsymbol {h}_1},\ldots,{\boldsymbol {h}_{q - 1}}) \in {\mathbb{R}^q}$, level order $\boldsymbol{q} \in {\mathbb{N}^q}$.
\Ensure $\boldsymbol K_{\boldsymbol{q}}$.
\State Calculate ${\boldsymbol t_{\boldsymbol {i}}} = \kappa\left({ \Vert [{{{\boldsymbol {i}}_s}{{\boldsymbol {h}}_s}:s \in [q]}] \Vert_2}\right)$, $\forall \boldsymbol {i} \in [\boldsymbol{q}]$.
\State Let
${\boldsymbol D_{\boldsymbol {i},s}} = \left\{ {\begin{array}{*{20}{c}}
   {\{ 0\},}  \\
   {\{ {\boldsymbol {i}_s},{\boldsymbol{q}_s} - {\boldsymbol {i}_s}\},}  \\

 \end{array} } \right.\begin{array}{*{20}{c}}
   {{\boldsymbol {i}_s} = 0,}  \\
   {1 \leqslant {\boldsymbol {i}_s} \leqslant {\boldsymbol{q}_s} - 1,}  \\

 \end{array}$  and ${\boldsymbol D_{\boldsymbol {i}}}={\boldsymbol D_{\boldsymbol {i},0}} \times {\boldsymbol D_{\boldsymbol {i},1}} \times  \cdots  \times {\boldsymbol D_{\boldsymbol {i},q - 1}}$, $\forall\boldsymbol {i} \in [\boldsymbol{q}]$ and $\forall s \in [q]$.
\State Calculate $\boldsymbol k_{\boldsymbol {i}} =\sum\limits_{\boldsymbol {j} \in {\boldsymbol D_{\boldsymbol {i}}}}{\boldsymbol t_{\boldsymbol {j}}}$.
\State \Return $\boldsymbol K_{\boldsymbol{q}} = circ_{\boldsymbol{q}}[\boldsymbol k]$.
\end{algorithmic}
\end{algorithm}

For $\boldsymbol K_{\boldsymbol{q}}$ generated by Algorithm \ref {alg:1}, only $O(n)$ is required to store it since we only need to store the first column. By Lemma \ref {lem:1}, $\boldsymbol K_{\boldsymbol{q}}\boldsymbol x$  is equivalent to implementing $(1/n)\boldsymbol \phi^{*} \operatorname{diag}(\boldsymbol \phi \boldsymbol k) \boldsymbol \phi \boldsymbol x$, which can be realized efficiently in $O(n \log n)$ using the mFFT. According to Theorem \ref {the0:2}, ${\boldsymbol K_{\boldsymbol{q}}^{ - 1}}\boldsymbol x$ is equivalent to implementing $(1/n)\boldsymbol \phi^{*} (\operatorname{diag}(\boldsymbol \phi \boldsymbol k))^{-1} \boldsymbol \phi \boldsymbol x$, which also can be realized efficiently in $O(n \log n)$. In addition to its advantages in computation and space storage, MCM approximation also does not require any sampling techniques. Next, we will design a fast and effective method to solve KLR based on MCM.

\section{Fast Newton Method Based on MCM Approximation}
\label{section:3}
In this section, we first simplify the Newton equation of KLR, then approximate the coefficient matrix of the simplified Newton equation as an MCM, finally propose a fast Newton method based on MCM approximation.
\subsection{Simplify the Newton equation}
\label{subsection:3.1}
KLR is a convex optimization problem \cite{boyd2004convex}, and the local optimal solution must be the global optimal solution. For convex optimization issues, Newton method with at least quadratic convergence can be used to solve them. The gradient and Hessian are obtained by differentiating \eqref {equ:3} with respect to $\boldsymbol \alpha$. The gradient is
\begin{equation}\label{equ:5}
\nabla F(\boldsymbol \alpha ) = \lambda \boldsymbol K\boldsymbol \alpha - \frac{1}{n}{\boldsymbol K}(\boldsymbol y - \boldsymbol p) ,
\end{equation}
and the Hessian of \eqref {equ:3} is
\begin{equation}\label{equ:6}
{\nabla ^2}F(\boldsymbol \alpha ) = \frac{1}{n}{\boldsymbol K^ \top }\boldsymbol \Lambda \boldsymbol K + \lambda \boldsymbol K,
\end{equation}
where $\boldsymbol \Lambda$ is a diagonal matrix with $\boldsymbol \Lambda_{ii} = {\boldsymbol p_i}(1 - {\boldsymbol p_i})$.
Based on the \eqref {equ:5} and \eqref {equ:6}, we need to solve the following Newton equation
\begin{align}\label{equ:7}
\left({\boldsymbol K^ \top }\boldsymbol\Lambda \boldsymbol K + n\lambda \boldsymbol K\right){\boldsymbol {d}} = {\boldsymbol K}(\boldsymbol y - \boldsymbol p - n\lambda \boldsymbol \alpha ),
\end{align}
to update the current solution.
Obviously, in order to compute the Newton direction $\boldsymbol {d}$, the computational complexity is $O({n^3})$ per iteration, which is prohibitively expensive for large-scale problems. In order to reduce the computational cost, we first simplify the Newton equation.
Since the kernel matrix is symmetric, we have
\begin{equation}\label{equ:9}
\boldsymbol K(\boldsymbol \Lambda \boldsymbol K + n\lambda \boldsymbol I){\boldsymbol {d}} = \boldsymbol K(\boldsymbol y - \boldsymbol p - n\lambda \boldsymbol \alpha ).
\end{equation}
If the kernel matrix $K$ is positive definite, we can simplify \eqref{equ:9} as
\begin{equation}\label{equ:10}
(\boldsymbol \Lambda \boldsymbol K + n\lambda \boldsymbol I){\boldsymbol {d}} = \boldsymbol y - \boldsymbol p - n\lambda \boldsymbol \alpha.
\end{equation}
If the kernel matrix  $\boldsymbol  K$ is positive semidefinite, then the solution to \eqref {equ:9} is not necessarily unique, but the unique solution of \eqref {equ:10} is the solution of \eqref {equ:9}. Therefore, we can use the solution of \eqref {equ:10} as the Newton direction.

Replacing $\boldsymbol K$ in \eqref{equ:10} with $\boldsymbol K_{\boldsymbol{q}}$  generated by Algorithm \ref {alg:1}, we can further obtain the following approximated Newton equation
\begin{equation}\label{equ:11}
(\boldsymbol \Lambda \boldsymbol K_{\boldsymbol{q}} + n\lambda \boldsymbol I){\bar {\boldsymbol {d}}} = \boldsymbol y - \bar{\boldsymbol p} - n\lambda \boldsymbol \alpha,
\end{equation}
where $\bar {\boldsymbol p}_{i}=1/({{1 + {e^{ - \boldsymbol K_{\boldsymbol{q}}(i, \cdot)\boldsymbol \alpha }}}}$) and $\boldsymbol K_{\boldsymbol{q}}(i, \cdot)$ is the
$i$-th row of $\boldsymbol K_{\boldsymbol{q}}$.

If solving \eqref{equ:11} by the conjugate gradient method \cite{hestenes1952methods} to obtain an approximated Newton direction, the time complexity of each iteration is $O(n\log n)$ due to the use of mFFT.

In this case, the most efficient scheme to solve \eqref{equ:11} is to run conjugate gradient method $T$ loops with the computational complexity $O(Tn\log n)$. Since $T=O(n)$ for conjugate gradient method \cite{hestenes1952methods}, this scheme is still insufferable for large-scale problems. Then we expect to find a more efficient way to calculate the Newton direction.
\subsection{Approximate the Coefficient Matrix of Newton Equation using MCM}
\label{subsection:3.2}
In this section, we  approximate the coefficient matrix of equation \eqref{equ:11} as an MCM, then we can calculate the Newton direction more efficiently.

According to Theorem \ref {the0:2}, if we can approximate $\boldsymbol \Lambda \boldsymbol K_{\boldsymbol{q}} + n\lambda \boldsymbol I$ with an MCM, then we can directly calculate ${\bar {\boldsymbol {d}}}$ in $O(n\log n)$ time by using mFFT. Obviously, we only need to approximate $\boldsymbol \Lambda \boldsymbol K_{\boldsymbol{q}}$  since $n\lambda \boldsymbol I$ is already an MCM. To this end, we solve the least squared problem
\begin{equation}\label{equ:12}
\min_{\boldsymbol A_{\boldsymbol{q}} \in \mathbb{A}_{\boldsymbol{q}}}  \quad  \|\boldsymbol \Lambda \boldsymbol K_{\boldsymbol{q}}-\boldsymbol A_{\boldsymbol{q}}\|_F^2,
\end{equation}
where $\mathbb{A}_{\boldsymbol{q}}$ is the set of MCM of level order $\boldsymbol{q}$. Here we use the Frobenius Norm for simplicity, and the other norms of matrix can be used.

By working out the optimality condition of the problem \eqref {equ:12}, we obtain the following proposition.
\begin{proposition}\label{propo:1}
The optimal solution of the problem \eqref {equ:12} is
\begin{equation}\label{equ:13}
\boldsymbol A_{\boldsymbol{q}}=\tau \boldsymbol K_{\boldsymbol{q}},
\end{equation}
where $\tau = (1/n)\sum_{i=1}^{n} {\boldsymbol \Lambda_{ii}}$.
\begin{proof}
Let $\boldsymbol k$ be the first column of $\boldsymbol K_{\boldsymbol{q}}$ and $\boldsymbol a$ be the first column of $\boldsymbol A_{\boldsymbol{q}}$, where $\boldsymbol k=(\boldsymbol k_1, \boldsymbol k_2,\ldots, \boldsymbol k_n)$ and $\boldsymbol a=(\boldsymbol a_1, \boldsymbol a_2,\ldots, \boldsymbol a_n)$. According to the built-in periodicity of $\boldsymbol K_{\boldsymbol{q}}$ and $\boldsymbol A_{\boldsymbol{q}}$, the problem \eqref {equ:12} is equivalent to
\begin{equation}\label{equ:14}
\min_{a_{j} \in \mathbb{R}} \quad \sum_{i=1}^{n}\sum_{j=1}^{n}{(\boldsymbol \Lambda_{ii}\boldsymbol k_j-\boldsymbol a_{j})^2}.
\end{equation}
The first-order optimality conditions of the problem \eqref {equ:14} is
\begin{equation*}
\sum_{i=1}^{n}{(\boldsymbol \Lambda_{ii}\boldsymbol k_j-\boldsymbol a_{j})}=0, j=1,2,\ldots,n.
\end{equation*}
Hence, the optimal solution of the problem \eqref {equ:14} is
\begin{equation*}
\boldsymbol a_{j}=\frac{1}{n} \sum_{i=1}^{n} {\boldsymbol \Lambda_{ii}}\boldsymbol k_j,j=1,2,\ldots,n.
\end{equation*}
Thus the optimal solution of the problem \eqref {equ:12} is \eqref{equ:13}, which proves the proposition.
\end{proof}
\end{proposition}

According to Proposition \ref {propo:1}, we replace $\boldsymbol \Lambda \boldsymbol K_{\boldsymbol{q}}$ in \eqref {equ:11} with $\boldsymbol A_{\boldsymbol{q}}=\tau \boldsymbol K_{\boldsymbol{q}}$, and get the following equation
\begin{equation}\label{equ:15}
(\tau \boldsymbol K_{\boldsymbol{q}} + n\lambda \boldsymbol I){\tilde{\boldsymbol {d}}} = \boldsymbol y - \tilde{\boldsymbol p} - n\lambda \boldsymbol \alpha,
\end{equation}
where $\tilde {\boldsymbol p}_{i}=1/({{1 + {e^{ - \boldsymbol K_{\boldsymbol{q}}(i, \cdot)\boldsymbol \alpha }}}})$.

Then, we can obtain the following approximate Newton direction
\begin{equation}\label{equ:16}
\tilde{\boldsymbol {d}} =\frac{1}{\tau} {({\boldsymbol K_{\boldsymbol{q}}} + \tilde \tau \boldsymbol I)^{ - 1}} (\boldsymbol y -\tilde{\boldsymbol p} - n\lambda \boldsymbol \alpha),
\end{equation}
where $\tilde \tau={n\lambda}/{\tau}$.

According to the nice properties of MCM, the cost of  calculating the approximated  Newton direction \eqref {equ:16} is $O(n\log n)$, which is much less than $O(n^2\log n)$ and $O(n^3)$.

\begin{figure*}[ht]
\centering
\includegraphics[scale=0.8]{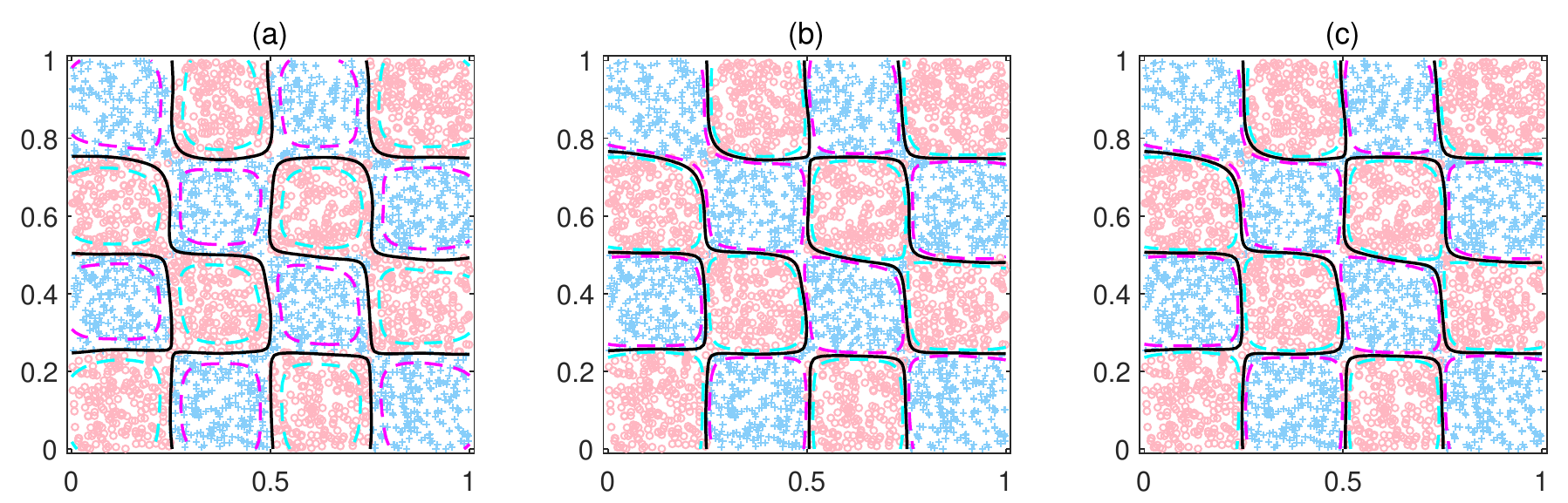}
\caption{Nonlinear classification experiments. $(a)$ Newton method based equation \eqref {equ:10}. $(b)$ Newton method based equation \eqref {equ:11}. $(c)$  Newton method based equation \eqref {equ:16}. The training samples of class $1$ are plotted as lightskyblue plus ($\color{lightskyblue}{+}$), and the training samples of class $0$ are plotted as lightpink circle ($\color{lightpink}{\circ}$). The solid black lines are the classification boundaries; the cyan dashed line and the magenta dashed line are the lines with predicted probabilities of 0.25 and 0.75, respectively. The test accuracies for the corresponding methods are 0.9585, 0.9584 and 0.9584 on 625 test samples, and the training time for the corresponding methods are 109.63s,  7.49s and 0.05s on 3375 train samples.}\label{fig:1}
\end{figure*}

To illustrate the effectiveness of using the approximated Newton direction \eqref {equ:16}, we compare the fast Newton method based on MCM approximation with the Newton method based on \eqref {equ:10} and the Newton method based on \eqref {equ:11} experimentally. A $2$-dimensional separable dataset for nonlinear classification was randomly sampled, which comprised 3375 training samples and 625 test samples. The training samples of class $1$ are plotted as lightskyblue plus ($\color{lightskyblue}{+}$), and the training samples of class $0$ are plotted as lightpink circle ($\color{lightpink}{\circ}$). The solid black lines are the classification boundaries; the cyan dashed line and the magenta dashed line are the lines with predicted probabilities of 0.25 and 0.75, respectively (Fig. \ref{fig:1}).

In Fig. \ref{fig:1}, it can be seen that the classification boundaries of the three Newton methods are almost the same. From the contour line of the predicted probability, the first method has more samples with the predicted probability between 0.25 and 0.75 than the latter two methods. The test accuracies for the corresponding methods are 0.9585, 0.9584 and 0.9584 on 625 test samples, and the training time for the corresponding methods are  109.63s,  7.49s and 0.05s on 3375 train samples. This validates the efficiency of the fast Newton method based on the MCM approximation.
\subsection{Fast Newton Method}
\label{subsection:3.3}
We are now ready to develop a fast Newton method based on MCM approximation, which can reduce the time and space complexity more succinctly and effectively.

The main work of Newton method is the calculation of Newton direction. According to Theorem \ref {the0:1},  $\boldsymbol K_{\boldsymbol{q}} + \tilde \tau \boldsymbol I$ is an MCM. By Theorem \ref {the0:2}, we have
\begin{equation*}
{({\boldsymbol K_{\boldsymbol{q}}} + \tilde \tau \boldsymbol I)^{ - 1}} = \frac{1}{n}{\boldsymbol \phi ^*}(diag(\boldsymbol v + \tilde \tau \boldsymbol {1}))^{-1}\boldsymbol \phi,
\end{equation*}
where $\boldsymbol v=\boldsymbol \phi\boldsymbol k$ is the vector of eigenvalues.

Then we rewrite \eqref {equ:16} as follows:
\begin{equation}\label{equ:17}
\tilde {\boldsymbol {d}}\!=\!\frac{1}{n\tau}{\boldsymbol \phi ^*}(diag(\boldsymbol v + \tilde \tau \boldsymbol {1}))^{-1}\boldsymbol \phi (\boldsymbol y\!-\!\tilde {\boldsymbol p}\!-\!n\lambda \boldsymbol \alpha ).
\end{equation}

In addition, replacing $\boldsymbol  K$ with $\boldsymbol K_{\boldsymbol{q}}$  generated by Algorithm \ref {alg:1}, we note the approximations of the objective function \eqref {equ:3} and its gradient \eqref {equ:5} as follows.
\begin{align}
\tilde F(\boldsymbol \alpha )\!&= \frac{\lambda }{2}\!{\boldsymbol \alpha ^ \top \!}{\boldsymbol K\!_{\boldsymbol{q}}\!}\boldsymbol \alpha\!-\!\frac{1}{n}\!({\boldsymbol y^ \top\! } \ln \tilde {\boldsymbol p}\! + {(\boldsymbol 1 - \boldsymbol y)\!^ \top \!}\ln (\boldsymbol 1\! - \tilde {\boldsymbol p}\!)) ,\label{equ:18}\\
\nabla \tilde F(\boldsymbol \alpha ) &={\boldsymbol K_{\boldsymbol{q}}}(\lambda \boldsymbol \alpha - \frac{1}{n}(\boldsymbol y - \tilde {\boldsymbol p})  ).\label{equ:19}
\end{align}
Now we present the detailed flow of the Fast Newton method based on MCM approximation in Algorithm \ref {alg:2}.
\begin{algorithm}[htp]
\caption{Fast Newton method based on MCM} \label{alg:2}
\begin{algorithmic}[1]
\Require Training set $\mathbb{D}$, parameters $\sigma$, $\lambda$, ${T}$, $\varepsilon$, $\delta\in (0,1)$, $\beta\in (0,0.5)$ and given an initial $\boldsymbol \alpha_0$.
\Ensure $\boldsymbol \alpha$.
\State Calculate $[\boldsymbol k_{\boldsymbol {i}}:\boldsymbol {i} \in [\boldsymbol{q}]]$ according to Algorithm \ref {alg:1}.
\State Calculate $\boldsymbol v=\boldsymbol \phi[\boldsymbol k_{\boldsymbol {i}}:\boldsymbol {i} \in [\boldsymbol{q}]]$ by mFFT.
\State Calculate $\tilde {\boldsymbol p}$, $\nabla \tilde F({\boldsymbol \alpha _0})$ and $\tilde F({\boldsymbol \alpha _0})$.
\While{$t \leqslant T_{max} $ and $|| {\nabla \tilde F({\boldsymbol \alpha _t})} || > \varepsilon $ }
    \State Calculate $\boldsymbol \eta  = (1/\tau)(\boldsymbol y - \tilde {\boldsymbol p} - n\lambda {\boldsymbol \alpha _t})$.
    \State Calculate $\boldsymbol \delta  = \boldsymbol \phi \boldsymbol \eta$ by mFFT.
    \State Calculate $\boldsymbol \zeta  = (diag(\boldsymbol v + \tilde \tau \boldsymbol {1}))^{-1}\boldsymbol \delta$.
    \State Calculate $\tilde {\boldsymbol {d}}= \frac{1}{n}{\boldsymbol \phi ^*}\boldsymbol \zeta$ using inverse mFFT.
    \State (Armijo line search \cite{jorge2006numerical}) Set $r_t=\delta^{m_t}$, where $m_t$ is the first nonnegative integer $m$ for which

     $\tilde F({\boldsymbol \alpha _t}+\delta^{m}\tilde {\boldsymbol {d}}) \leq \tilde F({\boldsymbol \alpha _t})+\beta {\delta^{m}}\langle\nabla \tilde F({\boldsymbol \alpha _t}),\tilde {\boldsymbol {d}}\rangle$.
    \State Update ${\boldsymbol \alpha _{t + 1}}{\text{ = }}{\boldsymbol \alpha _t}+r_t\tilde {\boldsymbol {d}}$, calculate $\tilde {\boldsymbol p}$ and $\nabla \tilde F({\boldsymbol \alpha _{t + 1}})$.
    \State $t:=t+1$.
\EndWhile
\State \Return $\boldsymbol \alpha  \leftarrow {\boldsymbol \alpha _t}$.
\end{algorithmic}
\end{algorithm}
\subsection{Complexity Analysis}
\label{subsection:3.4}
Due to the nice properties of MCM, our algorithm can reduce the time complexity and space complexity  in solving KLR more succinctly. In the following, we will study the time and space complexity of the Algorithm \ref {alg:2} in more detail.

\textbf{Space Complexity.} Because of the built-in periodicity of MCM, only $O(n)$ space storage is required. This plays an important role in the consumption of memory. Considering the fragmentation memory footprint of the other parameters, we can eventually abbreviate the space complexity of the Algorithm \ref {alg:2} to $O(n)$.

\textbf{Time Complexity.} From \cite{edwards2013approximate}, the complexity of the step 1 is $O(n)$. We know the computational complexity of steps 2, 6 and 8 is $O(n\log n)$, since mFFT can be applied with the $O(n\log n)$  complexity. The main work of the steps 3 and 10 is to calculate the form $K_{\boldsymbol{q}}\boldsymbol x$. According to Section \ref{subsection:2.2}, the computational complexity of the steps 3 and 10 is $O(n\log n)$.  The computational complexity of the step 7 is $O(n)$. In the process of calculating $\tilde F({\boldsymbol \alpha _t}+{\delta^{m}}\tilde {\boldsymbol {d}})$ in the step 9, we can store $\boldsymbol K_{\boldsymbol{q}}{\boldsymbol \alpha _t}$ and $\boldsymbol K_{\boldsymbol{q}} \tilde {\boldsymbol {d}}$  to facilitate the calculation of $\tilde F({\boldsymbol \alpha _t}+{\delta^{m}}\tilde {\boldsymbol {d}})$. Then only $O(n)$ times multiplications are needed to calculate $\tilde F({\boldsymbol \alpha _t}+{\delta^{m}}\tilde {\boldsymbol {d}})$ in the step 9. If the number of iterations is $T\leq T_{max}$, the total maximum computational complexity of Algorithm \ref {alg:2} is $O(nq+ Tn\log n)$. At the same time, \cite {ding2020approximate} showed that a small $q$ (e.g., 3) is sufficient for a sufficient approximation of the classification problem. This means that we can abbreviate  the time complexity of the Algorithm \ref {alg:2} to $O(Tn\log n)$, where $T$ is always less than 10 for the convex KLR problem.

We list the computational complexity and space complexity of several kernel approximation methods for solving KLR in Table \ref{table:1}.
\begin{table}[!htb]
\centering
\caption{Compare the typical approximation methods of solving KLR by Newton method. The space complexity of every method is in the second column. The time complexity per iteration for each method is in the third column. $n$ denotes the number of training data. $d$ denotes the data dimension. $c$ represents the sampling size. $D$ denotes the dimensionality of the random feature space.} \label{table:1}
\setlength{\tabcolsep}{1mm}
{
\begin{tabular}{lllllllll}
  \toprule
 Method          &Space complexity  & Time complexity\\
  \midrule
  Original         & $O(n^2)$      & $O(n^3+n^2)$ \\
  Nys \cite{NIPS2000_19de10ad}        & $O(nc)$       & $O(nc^{2}+nc+c^{3})$    \\
  RRLS-Nys  \cite{NIPS2015_f3f27a32}   & $O(nc)$       & $O(nc^{2}+nc+c^{3})$  \\
  SCRF \cite{feng2015random}       & $O(nD)$       & $O(nD^{2}+nD \log d+D^{3})$ \\
 LS-RFF \cite{Zhuli2021towards}       & $O(nD)$       & $O(nD^{2}+nD+D^{3})$ \\
Ours  & $O(n)$        & $O(n\log n)$    \\
  \bottomrule
\end{tabular}
}
\end{table}
\section{Experiment}
\label{section:4}
In this section, we conduct experiments on some binary and multi-classification datasets to evaluate the effectiveness of our algorithm. All the experiments were in MATLAB and run on a 3.6 GHz Intel Core i7 with 16GB of memory.
\subsection{Compared Methods and Parameter Settings}
\label{subsection:4.1}
We compare our method with the following state-of-the-art kernel matrix approximation methods:
\begin{itemize}
  \item   Nys \cite{NIPS2000_19de10ad}:  The standard form of Nystr{\"o}m method, whose sampling method uses uniform sampling.
  \item   RRLS-Nys  \cite{NIPS2015_f3f27a32}\footnote {Codes are available in \url{https://github.com/cnmusco/recursive-nystrom}.} : Recursive ridge leverage scores finds a more accurate kernel approximation in less time by employing a fast recursive sampling scheme.
  \item   SCRF \cite{feng2015random}: The transformation matrix is constructed by a signed Circulant Random Matrix (CRM) and the feature mapping can be done in $O(nD\log d)$  time by using the fast Fourier Transform (FFT).
  \item   LS-RFF \cite{Zhuli2021towards}\footnote {Codes are available in \url{http://www.lfhsgre.org}.}:  The transformation matrix is constructed based on the leverage score, which takes $O(nD^2+D^3)$ time to generate refined random features.
\end{itemize}

We fixed $T_{max} = 30$, and the stop criterion $\varepsilon  = {10^{ - 5}}$. The kernel function we use is $\kappa ({\boldsymbol x},{\boldsymbol z}) = \exp ( - \sigma {\Vert {{\boldsymbol x} - {\boldsymbol z}} \Vert^2})$, where $\sigma  > 0$ is the kernel parameter. For all the experiments, there are two parameters need to be determined in advance, i.e., $\sigma$ and  $\lambda$. The regularization parameter $\lambda$  and the kernel parameter $\sigma$  listed in Table \ref{table:2} were chosen by a cross-validation procedure and grid search with $\sigma  \in \{ {2^{ - 9}},\ldots,{2^{16}}\}$  and $\lambda  \in \{ {10^{ - 6}},\ldots,{10^0}\}$.

\begin{table}[!htb]
\centering
\caption{Parameter settings of algorithms.} \label{table:2}
\setlength{\tabcolsep}{0.3mm}
{
\begin{tabular}{llllll|lll}
  \toprule
  \multicolumn{6}{c|}{Binary} & \multicolumn{3}{c}{Multi-class}\\
\cline{1-6}   \cline{7-9}  
 Data Sets          &$\sigma$   & $\lambda$  & Data Sets      &$\sigma$     & $\lambda$  & Data Sets      &$\sigma$     & $\lambda$\\
  \hline

  Ionosphere       & $2^{2}$   & $10^{-3}$    & Adult      & $2^{-7}$     &$10^{-3}$  & Shuttle      & $2^{15}$     &$10^{-3}$\\
  Australian       & $2^{-7}$  & $10^{-2}$    & Shuttle    & $2^{15}$     & $10^{-3}$  & Sensorless    & $2^{9}$     & $10^{-4}$\\
  Banknote        & $2^{2}$   & $10^{-2}$   & Mnist     & $2^{-1}$      & $10^{-5}$   & Connect-4    & $2^{2}$     & $10^{-4}$\\
  Titanic         & $2^{-2}$  & $10^{-1}$   & Vehicle     & $2^{-4}$      & $10^{-1}$   & Mnist    & $2^{-1}$     & $10^{-6}$\\
  Banana          & $2^{3}$   & $10^{-3}$   & Skin        & $2^{11}$     & $10^{-3}$   & Vehicle    & $2^{5}$     & $10^{-3}$\\
  USPS         & $2^{-1}$  & $10^{-4}$   & Covtype     & $2^{9}$       & $10^{-4}$    & CovtypeM    & $2^{9}$     & $10^{-4}$\\
\bottomrule
\end{tabular}
}
\end{table}
The sampling size $c$  of the Nystr{\"o}m method and the recursive RLS-Nystr{\"o}m was set to $c = \sqrt n$. The dimensionality $D$  of SCRF and LS-RFF was set to $D{\text{ = }}O(d)$. For MCM approximation, a 3-level circulant matrix was adopted. In the experiments, we fix $\boldsymbol h=\boldsymbol 1 \in {\mathbb{R}^q}$, since it is sufficient to demonstrate the validity of the Algorithm \ref {alg:2}. To avoid randomness, all experiments are operated 10 times independently, and the mean value is taken  as the final result.

\subsection{Small-scale Benchmark Datasets Experiments}
\label{subsection:4.2}
In this section, we test six small-scale benchmark classification datasets to illustrate that the classification performance of the approximation algorithms is comparable to the original algorithm. For USPS, the task of classifying the digit 8 versus the rest classes was trained. The Australian and USPS datasets were downloaded from the LIBSVM \cite{libsvm2011}, and the Ionosphere, Banknote, Titanic and Banana datasets were downloaded from the UCI database \cite{Dua2019}. These datasets are detailed in Table \ref{table:3}.

\begin{figure}[!htb]
\centering
\includegraphics[scale=0.6]{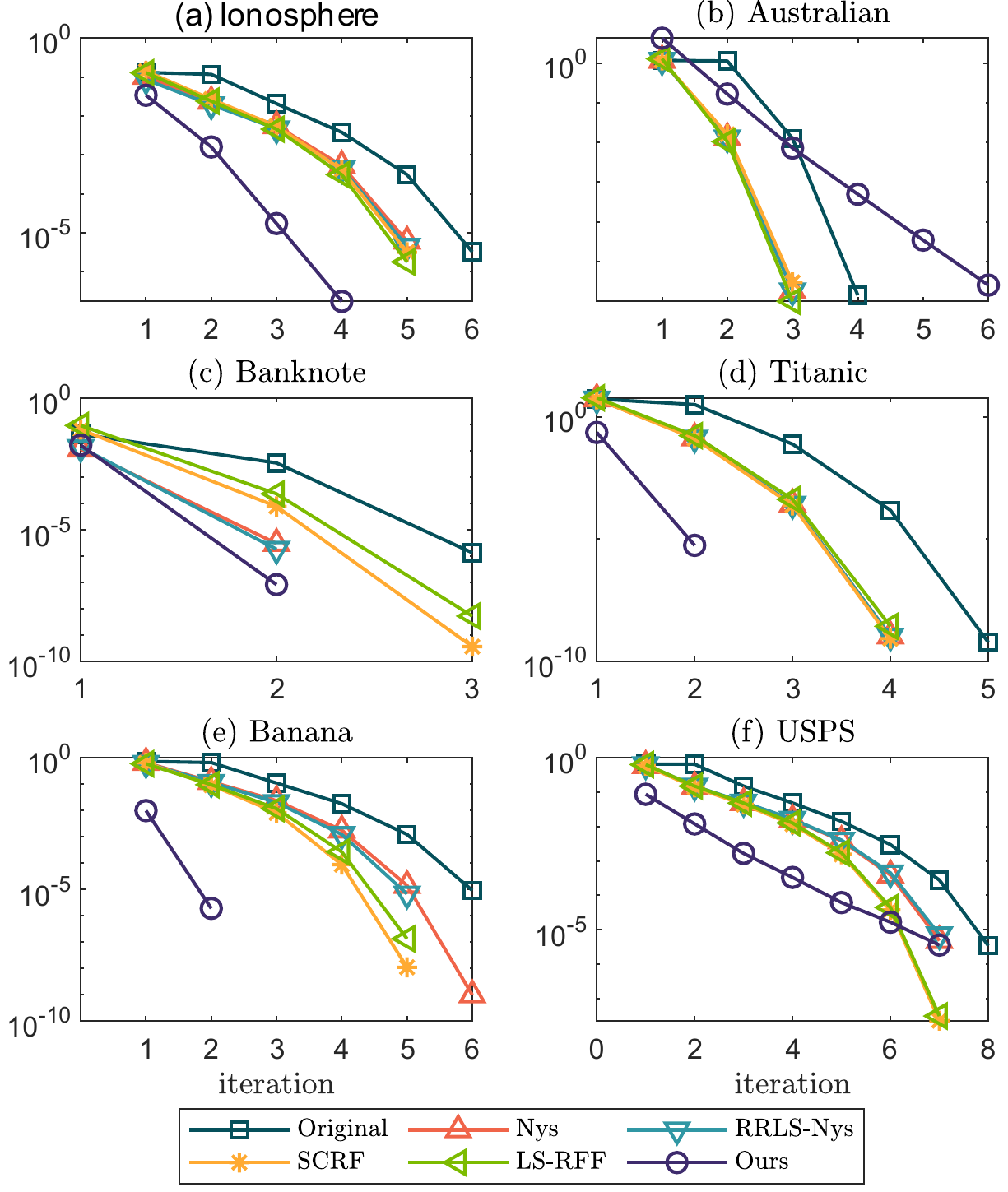}
\caption{Plots for the variation trend of the two-norm of the gradient with the number of iterations on the six small-scale benchmark classification datasets.}\label{fig:2}
\end{figure}

The variation trend of the two-norm of the gradient with the number of iterations is plotted in Fig. \ref{fig:2}. Table \ref{table:3} reports the experimental results for the six small-scale datasets. The training time is not listed in Table \ref{table:3}, because the training time of these methods is very small.

It can be seen from Fig. \ref{fig:2} that, compared with the original algorithm, several approximation algorithms can converge, and at the same time, the number of iterations when convergence is achieved will be reduced.
\begin{table}[!htb]
\centering
\caption{Comparison of different algorithms on the six small-scale benchmark classification datasets. The standard deviations are given in brackets. 'AUC' stands for "Area under the ROC Curve". $n$ and $m$ are the numbers of training and testing samples respectively. $d$ is the dimension of data.}\label{table:3}
\setlength{\tabcolsep}{2mm}
{
\begin{tabular}{llll}
\toprule
Data Sets       & Algorithms & Acc($\%$) & AUC($\%$)          \\
  \midrule
Ionosphere & Original   & 92.30        & 97.53(1.06) \\
n=216      & Nys        & 91.41        & 96.76(1.39) \\
m=135      & RRLS-Nys   & 91.41        & 97.71(1.13) \\
d=34       & SCRF       & 91.04         & 97.54(1.11) \\
           & LS-RFF       & 91.85         & 97.81(0.73) \\
           & Ours        & 92.22        & 97.81(1.00) \\
\midrule
Australian & Original   & 85.11        & 92.74(2.16) \\
n=512      & Nys        & 84.89        & 91.89(1.80) \\
m=178      & RRLS-Nys   & 84.38        & 92.38(1.69) \\
d=14       & SCRF       & 84.27        & 92.22(1.79) \\
           & LS-RFF       & 84.61        & 92.02(1.53) \\
           & Ours        & 87.64        & 92.54(2.19) \\
\midrule
Banknote   & Original   & 1.00          & 1.00(0.00) \\
n=1000     & Nys        & 99.95        & 1.00(0.00) \\
m=372      & RRLS-Nys   & 99.95        & 1.00(0.00) \\
d=4        & SCRF       & 99.87        & 1.00(0.00) \\
           & LS-RFF       & 99.92         & 1.00(0.00) \\
           & Ours        & 99.97        & 1.00(0.00) \\
\midrule
Titanic    & Original   & 77.41        & 74.75(1.00) \\
n=1331     & Nys        & 77.48        & 74.06(1.47) \\
m=870      & RRLS-Nys   & 77.51        & 74.50(1.65) \\
d=3        & SCRF       & 77.15        & 73.65(1.99) \\
           & LS-RFF     & 77.64         & 74.02(1.60) \\
           & Ours        & 77.31        & 73.58(1.05) \\
\midrule
Banana     & Original   & 90.54        & 96.82(0.18) \\
n=3430     & Nys        & 89.86        & 95.67(0.86) \\
m=1870     & RRLS-Nys   & 89.78        & 95.98(0.35) \\
d=2        & SCRF       & 87.46        & 93.62(1.56) \\
           & LS-RFF     & 86.65         & 93.34(1.47) \\
           & Ours        & 90.43        & 96.06(0.30) \\
\midrule
USPS       & Original   & 99.26        & 98.95(0.14) \\
n=7291     & Nys        & 99.25        & 98.83(0.24) \\
m=2007     & RRLS-Nys   & 99.26        & 98.90(0.18) \\
d=256      & SCRF       & 99.23        & 98.89(0.21) \\
           & LS-RFF       & 99.25         & 98.90(0.25) \\
           & Ours        & 99.13        & 99.33(0.14) \\
\bottomrule
\end{tabular}
}
\end{table}

From Table \ref{table:3}, it is clear that the classification performance of the five approximation algorithms is comparable to that of the original algorithm.
\subsection{Large-scale Benchmark Datasets Experiments}
\label{subsection:4.3}
In this section, we test six large datasets of benchmark classification to further demonstrate the superiority of our algorithm. For Mnist, the task of classifying the digit 8 versus the rest classes was trained. For Vehicle, the task of classify class 3 from the rest was trained. These datasets were downloaded from the LIBSVM \cite{libsvm2011}.

\begin{figure}[htb]
\centering
\includegraphics[scale=0.6]{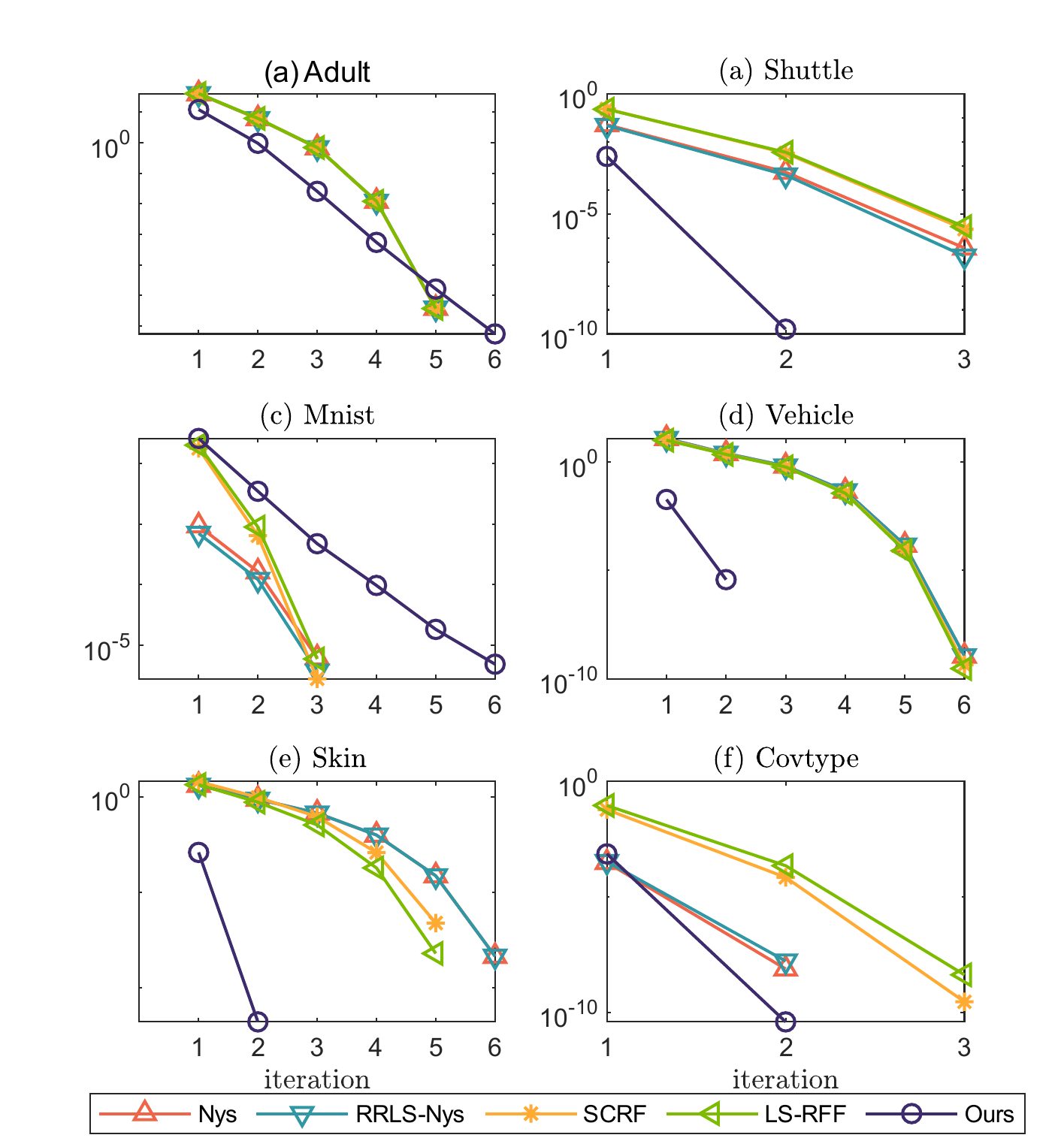}
\caption{Plots for the variation trend of the two-norm of the gradient with the number of iterations on the six large-scale benchmark classification datasets.}\label{fig:3}
\end{figure}

The variation trend of the two-norm of the gradient with the number of iterations is plotted in Fig. \ref{fig:3}. Table \ref{table:4} reports the experimental results for the six large-scale datasets.

From Fig. \ref{fig:3}, the five approximation algorithms can reach convergence in six large-scale data sets. The convergence process of Nys and RRLS-Nys is basically the same. The convergence process of SCRF and LS-RFF is basically the same. Several approximation methods can converge with very few iterations.
\renewcommand{\arraystretch}{1.01}
\begin{table}[!htb]
\centering
\caption{Comparison of different algorithms on the six large-scale benchmark classification datasets. 'AUC' stands for "Area under the ROC Curve". $n$ and $m$ are the numbers of training and testing samples respectively. $d$ is the dimension of data. The standard deviations are given in brackets.}\label{table:4}
\setlength{\tabcolsep}{1mm}
{
\begin{tabular}{lllll}
\toprule
Data Sets     & Algorithms & Training  & Acc & AUC            \\
               &            &  time(s)  &($\%$)    & ($\%$) \\
\midrule
Adult     & Nys        & 0.45(0.01)        & 81.79       & 88.39(0.00) \\
n=32,561  & RRLS-Nys   & 0.54(0.01)        & 81.82       & 88.39(0.01) \\
m=16,281  & SCRF       & 5.35(0.17)        & 81.72      & 88.40(0.12) \\
d=123     & LS-RFF       & 5.96(0.06)        & 82.37      & 88.22(0.10) \\
          & Ours        & 0.24(0.01)        & 82.08       & 86.55(0.02) \\
\midrule
Shuttle  & Nys        & 0.54(0.03)        & 99.85       &99.98(0.00) \\
n=43,500  & RRLS-Nys   & 0.89(0.07)        & 99.85       & 99.98(0.00) \\
m=14,500  & SCRF       &  0.25(0.05)         & 99.84       & 99.97(0.00) \\
d=9      & LS-RFF       & 0.33(0.05)        & 99.84      & 99.98(0.00) \\
          & Ours        & 0.32(0.02)        & 99.85       & 99.97(0.00) \\
\midrule
Mnist    & Nys        & 1.20(0.04)        & 99.31       & 99.47(0.01) \\
n=60,000  & RRLS-Nys   & 2.45(0.06)        & 99.32       & 99.47(0.01) \\
m=10,000  & SCRF       & 9.63(0.23)       & 99.29       & 99.46(0.00) \\
d=784   & LS-RFF       & 12.06(0.15)        & 99.30      & 99.47(0.00) \\
        & Ours      & 0.46(0.01)        & 99.39       & 99.62(0.01) \\
\midrule
Vehicle  & Nys        & 1.32(0.05)        & 85.37       & 91.44(0.00) \\
n=78,823  & RRLS-Nys   & 1.44(0.03)        & 85.38       & 91.45(0.00) \\
m=19,705  & SCRF       & 7.70(0.16)        & 83.95       & 90.50(0.00) \\
d=100   & LS-RFF       & 6.64(0.09)        & 83.95       & 90.50(0.09) \\
        & Ours        & 0.56(0.03)        & 83.92       & 87.11(0.03) \\
\midrule
Skin     & Nys        & 9.30(0.21)        & 99.94       & 99.96(0.00) \\
n=157,464 & RRLS-Nys   & 9.20(0.90)        & 99.94       & 99.96(0.00) \\
m=87,593  & SCRF       & 0.46(0.01)        & 99.93       & 99.95(0.00) \\
d=3     & LS-RFF       & 0.56(0.57)        & 99.93       & 99.95(0.00) \\
        & Ours       & 1.12(0.01)        & 99.94       & 99.97(0.00) \\
\midrule
Covtype  & Nys        & 15.96(0.17)       & 95.14       & 97.81(0.03) \\
n=456,533 & RRLS-Nys   & 13.85(0.58)       & 95.13       & 97.81(0.02) \\
m=124,479 & SCRF       & 4.85(0.39)       & 95.11       & 97.80(0.00) \\
d=54    & LS-RFF       & 6.52(0.11)        &95.15     & 97.80(0.00) \\
        & Ours        & 3.36(0.07)        & 95.15       & 97.81(0.03)\\
\bottomrule
\end{tabular}
}
\end{table}

From Table \ref{table:4}, we have four observations. First, our algorithm's classification performance is comparable to the other four algorithms on all datasets except Vehicle. For Vehicle, the other four algorithms have slightly higher accuracy and AUC than our algorithm, but our algorithm has the least training time. Second, the performance of SCRF is slightly worse than that of Nys and RRLS-Nys, and the performance of Nys and RRLS-Nys is similar, which is consistent with the conclusions of \cite {NIPS2017_a03fa308}. Third, the larger the dimension of training set is, the time cost of SCRF and LS-RFF increases obviously, which is consistent with the complexity analysis. Fourth, it can be clearly seen that the larger the training set is, the more obvious the efficiency gain of our algorithm is, which is consistent with the results of complexity analysis.
\subsection{Checkerboard Dataset Experiments}
\label{subsection:4.4}
In order to further illustrate the superiority of our algorithm, the influence of the sampling size of the Nys and RRLS-Nys on the classification performance of Checkerboard dataset is specifically analyzed in this section.

Checkerboard dataset was first proposed in \cite {ho1996building} and later widely used to illustrate the effectiveness of nonlinear kernel method \cite {mangasarian2001lagrangian,zhou2016sparse,chen2018sparse}. Checkerboard dataset was generated by the following method: randomly sampled 1,600,000 points from the regions $[0,1] \times [0,1]$ and labeled two classes by $4 \times 4$ XOR problem. Then we randomly chose 1,000,000 points as training samples and the remaining 600,000 points as test samples.

\begin{figure}[t]
\centering
\includegraphics[scale=0.5]{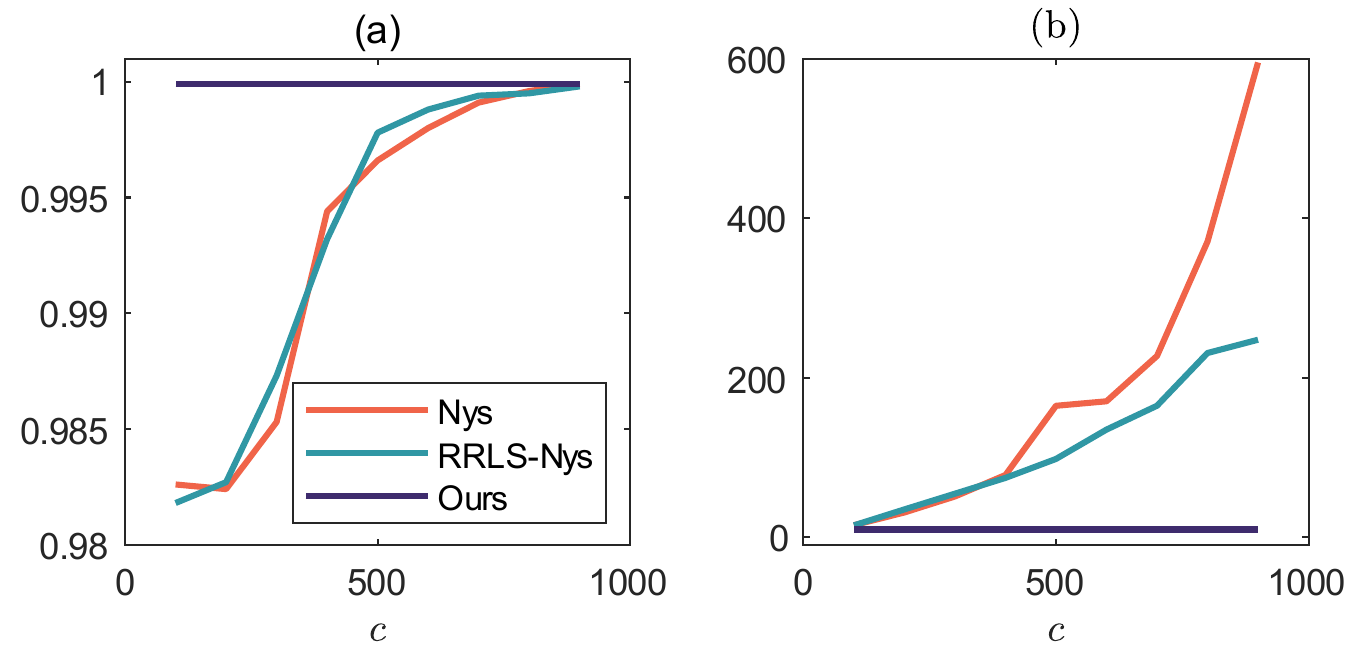}
\caption{AUC and training time with different sampling size on the Checkerboard dataset.}\label{fig:4}
\end{figure}

In order to analyze the influence of the sampling size on the classification performance of the Nys and RRLS-Nys, we gradually increased the sampling size from 100 to 900. Fig. \ref{fig:4} shows the variation of AUC and training time with the sampling size for Nys and RRLS-Nys, respectively.

As can be seen from Fig. \ref{fig:4}(a), the performance of the Nys and RRLS-Nys improves with the increase of the sampling size. However, the performance of the Nys and RRLS-Nys cannot be comparable to that of our method until the sampling size is 900. In addition, it can be seen from Fig. \ref{fig:4}(b) that the training time of Nys and RRLS-Nys increased significantly with the increase of the sampling size, which is much larger than ours. Based on the above analyses, we conclude that our method is more suitable for the Checkerboard dataset.
\subsection{Multi-class Classification}
\label{subsection:4.5}
In this section, Multi-classification experiments were performed by one-versus-all \cite {vapnik2013nature} on the six benchmark multi-classification datasets. In order to compare the performance of the five approximation algorithms objectively and impartially,  we also adopted the Macro averaged F1 scores (Macro-$F_1$) \cite {narasimhan2015consistent} and the Matthews correlation coefficient (MCC) \cite {gorodkin2004comparing} as the evaluation criteria. The detailed information of the six benchmark multi-classification datasets are listed in Table \ref{table:5}. And these datasets were downloaded from the LIBSVM \cite{libsvm2011}.
\renewcommand{\arraystretch}{1.1}
\begin{table}[!htb]
\centering
\caption{Datasets used in multi-classification experiments.}\label{table:5}
\setlength{\tabcolsep}{1mm}
{
\begin{tabular}{lllllll}
\toprule
Data Sets       & Train num. & Test num.    &Features &Classes  \\
\midrule
  Shuttle       & 43,500        & 14,500    & 9      & 7     \\
  Sensorless    & 46,656        & 11,853    & 48     & 11\\
  Connect-4     & 54,872        & 12,685    & 126    & 3    \\
  Mnist         & 60,000        & 10,000    & 784    & 10  \\
  Vehicle       & 78,823        & 19,705    & 100    & 3   \\
  CovtypeM      & 456,533       & 124,479   & 54     & 7 \\
\bottomrule
\end{tabular}
}
\end{table}

Table \ref{table:6} reports the experimental results for the six benchmark multi-classification datasets.
\renewcommand{\arraystretch}{1.1}
\begin{table}[!htb]
\centering
\caption{ Comparison of different algorithms on the six benchmark multi-classification datasets. The standard deviations are given in brackets.}\label{table:6}
\setlength{\tabcolsep}{0.8mm}
{
\begin{tabular}{llllll}
\toprule
Data      & Algorithms & Training & Acc & Macro-$F_1$   & MCC \\
  Sets             &  & time(s) & ($\%$) & ($\%$)    & ($\%$) \\
\midrule
\multirow{4}{0.5cm}{\rotatebox[origin=c]{90}{Shuttle}}     & Nys        & 5.78(0.10)        & 99.77       & 73.81(0.11)   & 99.35(0.01) \\
      & RRLS-Nys   & 8.70(0.37)        & 99.78       & 74.03(0.50)  & 99.35(0.01) \\
        & SCRF       & 1.81(0.05)       & 99.77       & 73.86(0.14) &99.35(0.00)\\
          & LS-RFF       & 2.27(0.09)       & 99.78       & 73.94(0.13) &99.36(0.00)\\
        & Ours         & 2.39(0.06)       & 99.77       & 73.76(0.00) &99.35(0.00)\\
\midrule
  \multirow{4}{0.5cm}{\rotatebox[origin=c]{90}{Sensorless}}  & Nys        & 9.13(0.06)        & 98.92       & 98.91(0.01)   & 98.81(0.11) \\
          & RRLS-Nys   & 9.81(0.18)        & 98.91       & 98.92(0.01)  & 98.81(0.11) \\
        & SCRF       & 15.19(0.15)       & 98.90      & 98.90(0.01) &98.79(0.01)\\
        & LS-RFF       & 19.37(0.17)       & 98.93      & 98.93(0.00) &98.82(0.00)\\
        & Ours        & 4.18(0.05)       & 98.91       & 98.91(0.01) &98.80(0.01)\\
\midrule
   \multirow{4}{0.5cm}{\rotatebox[origin=c]{90}{Connect-4}}  & Nys        & 1.53(0.03)        & 80.94       & 57.55(0.49)   & 58.63(0.54) \\
          & RRLS-Nys   & 2.15(0.05)        & 80.78       & 57.45(0.55)  & 58.50(0.73) \\
        & SCRF       & 8.60(0.07)       & 81.16       & 58.29(0.58) &59.13(0.65)\\
        & LS-RFF       & 10.90(0.09)       & 81.31       & 58.16(0.41) &59.30(0.44)\\
        & Ours       & 1.33(0.05)       & 81.71       & 59.68(0.59) &60.55(0.54)\\
\midrule
   \multirow{4}{0.5cm}{\rotatebox[origin=c]{90}{Mnist}}  & Nys        & 15.97(0.30)        & 96.80       & 96.79(0.00)   & 96.44(0.00) \\
          & RRLS-Nys   & 29.24(0.34)        & 96.80       & 96.79(0.03)  & 96.44(0.04) \\
        & SCRF       & 97.26(3.09)       & 96.74       & 96.74(0.01) &96.38(0.02)\\
             & LS-RFF       & 122.4(0.56)       & 96.80      & 96.80(0.41) &96.45(0.03)\\
        & Ours        & 8.37(0.16)       & 96.75      & 96.74(0.00) &96.39(0.00)\\
\midrule
   \multirow{4}{0.5cm}{\rotatebox[origin=c]{90}{Vehicle}}     & Nys        & 3.08(0.13)       & 82.77      & 82.00(0.01)   & 72.78(0.01) \\
          & RRLS-Nys   & 3.57(0.27)        & 82.80       & 82.02(0.01)  & 72.82(0.11) \\
        & SCRF       & 16.95(0.10)       & 82.77       & 81.98(0.01) &72.76(0.01)\\
        & LS-RFF       & 20.68(0.18)       & 82.86       & 82.09(0.00) &72.92(0.03)\\
        & Ours       & 1.77(0.05)       & 82.76       & 81.98(0.01) &72.76(0.01)\\
\midrule
   \multirow{4}{0.5cm}{\rotatebox[origin=c]{90}{CovtypeM}}  & Nys        & 110.8(1.85)       & 94.08      & 90.25(0.16)   & 90.49(0.01) \\
          & RRLS-Nys   & 92.95(1.85)        & 94.13       & 90.27(0.16)  & 90.56(0.12) \\
        & SCRF       & 101.1(0.45)       & 94.06     & 90.09(0.13) &90.45(0.01)\\
        & LS-RFF       & 130.4(1.37)       & 94.07      &90.10(0.33) &90.48(0.11)\\
        & Ours       & 24.21(0.41)       & 94.09       & 90.11(0.19) &90.51(0.01)\\
\bottomrule
\end{tabular}
}
\end{table}

From Table \ref{table:6}, it can be clearly seen that our method always has the least time cost, which is consistent with the time complexity. Judging from different evaluation criteria, the classification performance of the five approximation algorithms is neck and neck. Furthermore, it is clear that the larger the training set, the more significant the speedup of our method. This proves that our method is more effective to solve the multi-classification problem.
\section{Conclusion}
\label{section:5}
Kernel Logistic Regression (KLR) has a direct probabilistic interpretation and has good performance in many classification problems. However, the time and space complexity are prohibitive for large-scale issues. In this paper, we employ multilevel circulant matrix (MCM) approximation to save storage space and accelerate the solution of the KLR. Combined with the characteristics of MCM and our inspiring design, we propose a fast Newton method based on MCM approximation. Because MCM's built-in periodicity allows the multidimensional fast Fourier transform (mFFT) to be used in our method, the time complexity and space complexity of each iteration are reduced to $O(n \log n)$ and $O(n)$, respectively. The experimental results show that our method makes KLR scalable for some large-scale binary and multi-class problems. At the same time, our method provides faster speed and less memory consumption for training. In addition, the experimental results also show that the larger the training set size, the more significant the speedup of our method. Therefore, fast Newton method based on MCM approximation is a more suitable choice for handling large-scale KLR problems.

In this paper, KLR is taken as an example to study the application of MCM in kernel approximation. The kernel approximation method and its Newton equation approximation technique can be used in other kernel learning, such as support vector machines \cite{zhou2013new,chauhan2019problem,shuisheng2020unified}, to reduce the storage space and time complexity.

\bmhead{Acknowledgments}

This work was supported by the National Natural Science Foundation of China [Grants numbers 61772020].
\section*{Declarations}
\begin{itemize}
\item Data availability statement

Some or all data, models, or code generated or used during the study are available from the corresponding author by request.
\end{itemize}

\end{document}